\documentclass[copyright,creativecommons]{eptcs}
\providecommand{\event}{ICLP 2023} 

\usepackage{iftex}

\ifpdf
  \usepackage{underscore}         
  \usepackage[T1]{fontenc}        
\else
  \usepackage{breakurl}           
\fi

\usepackage{amsmath}
\usepackage{graphicx}
\usepackage{multirow}
\usepackage{amssymb}
\usepackage{url}

\title{An xAI Approach for Data-to-Text Processing with ASP\,\thanks{Partially supported by the \emph{Interdepartmental Project on Artificial Intelligence} (Strategic Plan UniUD--22-25) and by INdAM--GNCS projects CUP E55F22000270001 and CUP E53C22001930001.}}
\author{Alessandro Dal Pal{\`u}
\institute{Universit\`a di Parma, Italy}
\institute{GNCS-INdAM, Roma, Italy}
\email{alessandro.dalpalu@unipr.it}
\and
Agostino Dovier \qquad\qquad Andrea Formisano
\institute{Universit\`a di Udine, Italy}
\institute{GNCS-INdAM, Roma, Italy}
\email{\quad agostino.dovier@uniud.it \qquad andrea.formisano@uniud.it}
}
\def\titlerunning{An xAI Approach for Data-to-Text Processing with ASP}
\def\authorrunning{A. Dal Pal{\`u}, A. Dovier \& A. Formisano}
\begin{document}
\maketitle

\begin{abstract}
The generation of natural language text from data series gained renewed interest among AI research goals.
Not surprisingly, the few proposals in the state of the art are based on training some system, in order to produce a text that describes and that is coherent to the data provided as input.
Main challenges of such approaches are the proper identification of \emph{what} to say (the key descriptive elements to be addressed in the data) and \emph{how} to say:
the correspondence and accuracy between data and text, the presence of contradictions/redundancy in the text, the control of the amount of synthesis.

This paper presents a framework that is compliant with xAI requirements.
In particular we model ASP/Python programs that enable an explicit control of accuracy errors and amount of synthesis, with proven optimal solutions.
The text description is hierarchically organized, in a top-down structure where text is enriched with further details, according to logic rules.
	The generation of natural language descriptions' structure is also managed by logic rules.
\end{abstract}


\section{Introduction}

The last decades witnessed a remarkable attention to the impact of Artificial Intelligence (AI)
on many aspects of everyday life.
The ensuing debate underscores the social and ethical implications of the use of AI-based systems and
is maturing into new regulations in USA and in Europe~\cite{USA21,EU21}.
The European proposal should be operational by 2024 and it introduces a
risk-based classification of AI systems based on the principles of ethical AI~\cite{fgene.2022.927721}.

The term \emph{explainable AI} (xAI, see, for instance,~\cite{ArrietaRSBTBGGM20,AdadiB18})
has emerged to capture desirable properties of high-risk systems:
e.g., transparency (or glass box approach), ethics, the capability
to support results in terms of intelligible descriptions, accountability, security, privacy, and fairness.
In the near future, the adoption of AI systems will
depend on the capability of providing a high-level description of inner activities, which promotes the interpretability and
transparency of the decisions that lead to a result. Hence, domain experts could easily understand and criticize the contributions
coming from an AI system, in a similar way colleague’s opinions would be considered.

In general terms, the data-to-text generation task consists in automatically generating descriptions from non-linguistic data.
In recent years, there has been growing interest in such systems
(see for instance the work \cite{harkous-etal-2020-text}, supported by the Alexa AI organization).
Systems able to textually summarize data, such as time-series,
can help in making data more accessible to non-experts, compared to traditional forms of presentations.
Indeed, natural language descriptions of data can help readers even in cases where data are depicted by means of
forms of representations usually considered ``self-explanatory'', such as charts, histograms, pies, etc,
that might be difficult to read if the amount and/or complexity of data is large.

Text-to-speech tools, such as screen readers,  might exploit textual
generator to extract information from charts contained in a digital document.
This is also relevant in situations where the interpretation of visualizations 
is made difficult or hindered for people with visual impairments or when readers have 
limited cognitive abilities in comprehending and analyzing complex charts.

Variants of this problem have been widely studied and solutions have been proposed in various domains.
For example, \cite{Puduppully0L19} and \cite{WisemanSR17}
apply a data-to-text generation system to interpret data-records concerning NBA basketball games.
In the financial field, \cite{MurakamiWMGYTM17} proposes an approach to the problem of comment
generation for time-series of stock prices.
Different solutions have been proposed expressively to treat specific forms of diagrammatic representations of data
such as charts and histograms \cite{ObeidH20,abs-1812-10636} or tables~\cite{LiuWSCS18}.
In the healthcare domain, 
\cite{PortetRGHSFS09} presents the prototype BT-45 aimed at generating textual summaries of
physiological clinical data (acquired automatically from sensors or entered routinely by the medical staff).
For space limits, we cannot provide an exhaustive description of the literature concerning data-to-text generation.
We refer the reader to the previously mentioned papers and to references therein.

Almost all the recent approaches to data-to-text are based on Machine Learning (ML)
and ultimately rely on Deep Learning (DL) techniques, usually  exploiting Neural Networks training algorithms.
For example, the paper 
\cite{harkous-etal-2020-text}, mentioned earlier, proposes DataTuner a neural data-to-text generation system.

There are a few exceptions worth mentioning. Among them we mention the seminal work \cite{Kukich83}
that proposes a rule-based approach (actually, an expert system written in the  OPS5 production system language)
to generate descriptions of inputs from a Dow Jones stock quotes database.
Another notable exception is the BT-45 system cited earlier.
It exploits signal analysis and pattern detection techniques
to process input time series and relies on a corpus of human-authored sample  linguistic expression of concepts 
to be used in generating textual summaries.
The system is one of the few proposals appeared in the literature that do  not rely on ML/DL techniques.
In this sense, it is similar to our approach. Nevertheless, it is conceived and designed for a very specific application field.
Indeed, it requires the development, by human experts, of an ontology of medical concepts. Such an ontology is essential
both for assessing the importance of patterns in input data and for generating expert-oriented output descriptions (see \cite{PortetRGHSFS09} for a detailed presentation of the system).

\medskip

It should be noted that Machine Learning and Deep Learning methods optimize inner parameters (sub-symbolic representation)
to adhere to training targets, while they
struggle to produce a high-level description (or symbolic) about how and why the network learned specific
parameters (black box approach).
This limitation favoured post-hoc explanations \cite{ArrietaRSBTBGGM20},
 e.g., textual captioning and/or visual explanation by examples, still powered by an additional sub-symbolic machinery.
 Even though explanations in Deep Learning approaches are one of the next big challenges for AI,
 \cite{Rud19} expresses some concerns about the accuracy achievable by this type of explainability and postulates that interpretable
models for high-stakes decisions should be sub-symbolic free systems. 

The urge for explainability represents an opportunity of discontinuity with respect to current AI methods.
We divert from mainstream methods based on ML and DL, and the recent trend of producing a
high-level explanation of such black box systems.
The design of an architecture that is explainable and ML/DL free
represents an innovative and unexplored research direction.

Our strategic choice is to use robust and off-the-shelf technologies to reach xAI compliance.
We select Answer Set Programming (ASP) and Clingo~\cite{DBLP:series/synthesis/2012Gebser}  as the explainable core of the system, because they
allow knowledge representation at a high-level and enables reasoning about such knowledge.
This framework not only is explainable, but also domain experts, in a scientists-in-the-loop fashion,
can detect flaws in the automated process and review the knowledge base accordingly.
Handling of incoherent points of view, inconsistent rules, uncertainty and lack of knowledge can be explicitly supported by the system.

\section{System's Design}

Our systems models the selection of \emph{what} to say in (which descriptions are interesting to be narrated) and \emph{how} to say (how to build the qualitative features and how to create a narration structure) the final narration. The core of the system is explainable by design, since we develop a transparent deductive handling of semantic rules that describe features, by exploiting ASP. The descriptions selection is done through two optimization steps performed by two ASP programs which determine, possibly multiple, descriptions for each relevant region of the input. Then,
the best candidate descriptions (for different levels of detail) are selected.
The outcome of this selection phase is then processed, again by using ASP, to determine structural relations among the fragments (narration order, dependencies, etc) to be used for the generation of the narration. A Python script controls the pre-computation of input, the handling of ASP programs execution and the conversion of ASP's output to text and enriched plots. 

\subsection{Descriptions and curve fitting}
Let us present the overall idea and specific explainability targets we set in the process of conversion of a data series to text. In this work we focus on scalar data acquired on a time domain (represented by a sequence of floats). Each element can be associated to a timestamp. A description of the data is a simple and peculiar feature that generalizes a particular distribution along time. A possible bridge from data to robust descriptions is to rely on fitting functions. We select function prototypes that are fitted to data (i.e., least square curve fitting) and that are described by few and intuitive parameters, in order to promote a simple textual description. In the paper we refer to \emph{description/descriptor} as the function model and its parameters, while  we refer to \emph{narration} as the corresponding natural language wording of such description. 
As an example let us consider the concept of a valley in a series: a sudden decrease followed by another sudden increase of values can be modeled by a specific function (e.g., a V like function). The feature extraction, based of curve fitting, allows to retrieve which optimal parameters provide minimal Root Mean Square Error (RMSE) of the curve w.r.t.\ the original data. 
Parameters, that control the degrees of freedom of the shape of the curve, allow to build a high level narration. For example, if the two sides of the valley have a high angular coefficient (or vertical orientation), the valley can be described as sharp. Potential domain-dependent descriptors can be encoded (e.g., specific patterns) and added with no relevant modification of the system. 

In our model, we allow an arbitrary large number of functions prototypes, to support a modular and general usage. Each of them allows to investigate particular behaviors. Each prototype requires to encode its fitting function and its conversion to text as natural language description (we discuss about it later). In our system we currently encoded (number of parameters inside brackets): line (2), two segments polyline (4), a tooth like function (5) and the most prominent sinusoid (3). We plan for the future to include the analysis of transformed spaces (e.g., Fourier transform and wavelets) and patterns occurring with characteristic shapes (e.g., EEG tracks).

Such choice allows to accomplish some steps towards explainability. In particular, measuring the accuracy of a description, allows the system to provide a self-judgement feedback. There is a measurable error (the fitting accuracy) and a more qualitative error caused by conversion of descriptions (numerical) to narrations (textual).

\begin{figure}[ht]
 \centering
\includegraphics[width=0.85\linewidth]{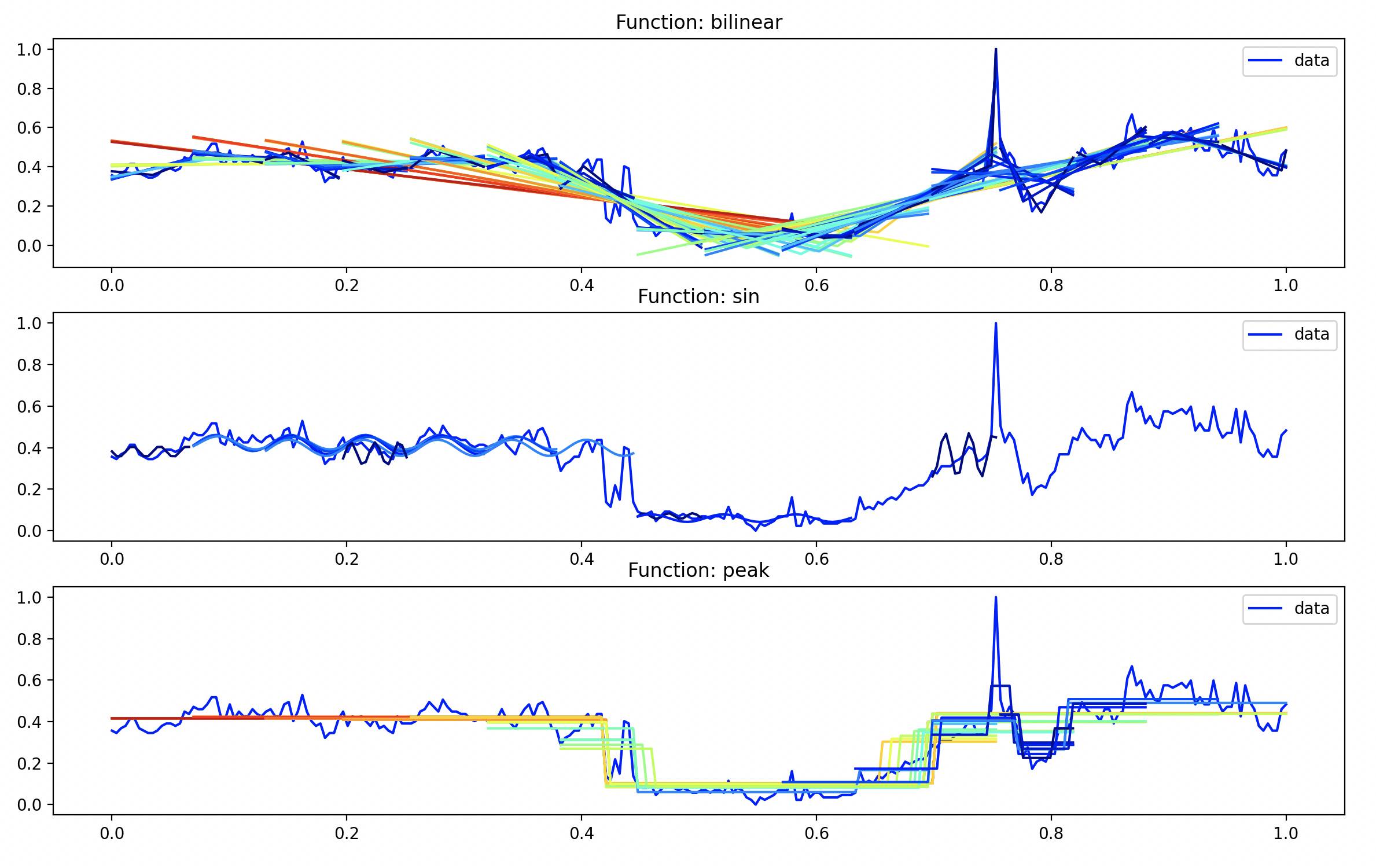}
\caption{Example of three function prototypes fitting for 16 zones}
\label{fig:fitting}
\end{figure}

When interpreting the evolution of a measure along time, often we extrapolate a multi-resolution description: in the simplest case we focus on the overall trend and also on remarkable details. Therefore, we pre-process descriptors at different sub-intervals, in order to capture potential features at different time scales.
Formally, we divide the time domain into regular $n=2^\ell$ zones. Each zone covers a subset of time samples and it is the smallest unit for RMSE optimizations. The parameter $\ell$ (levels of divisions) controls the amount of time resolution is provided to descriptors. Each descriptor $d$ covers a zone range from $i$ to $j$: $Z_d=\{z: 0 \leq i\leq z \leq j< n \}$. In Figure~\ref{fig:fitting} we show an example for some descriptors (bilinear, sinusoid and squared peak), colored by zones size, for a series resulting to the search term ``Concert'' in the last 5 years (from Google Trends) and normalized between 0 and 1.
Given a descriptor $d$, we measure RMSE for each zone $i\in Z_d$, namely $Err(d,i)$. The error is set to infinity for zones not in $Z_d$.

\subsection{Optimal descriptions covering}
The problem of finding optimal descriptions that minimize the overall RMSE error is rather trivial and it is the first phase in the overall optimization process. In a practical scenario, we can expect thousands of description candidates (set $\mathcal{D}$) to be processed. They originate from a rather small set of function prototypes combined with a large set of sub-intervals of $2^\ell$ to be investigated. 

The goal is to represent the series through a set of descriptions $D \subseteq \mathcal{D}$ out of the description pool. Each description $d$ has an associated set of zones $Z_d$, the curve type, optimal parameters and the RMSE $Err(d,i)$.

We wish to control the level of synthesis by an explicit parameter. In other DL approaches this is rather complex to handle, especially in case of rather long descriptions because of the incoherence in long text generation. We introduce the concept of \emph{verbosity} ($v=|D|$) that defines the maximal number of descriptions to be employed to describe the series. The verbosity corresponds to the number of sentences used in the final narration.

The problem can be stated as follows. Given a verbosity $v$, select a set $D_v$ such that $|D_v|=v$ and the sum of errors $Err(d,i)$ of each description $d\in D_v$ for each zone $i\in Z_d$ is minimal. We also ask for no zones overlaps ($Z_d \cap Z_d' = \emptyset$, for each $d,d'\in D_v$) and that each description covers at least $n/2^v$ zones. This last requirement ensures that for low verbosity the descriptions cover a sufficient time span, as a preference for general descriptions rather than extremely small details.

\begin{figure}[tb]
 \centering
\includegraphics[width=0.85\linewidth]{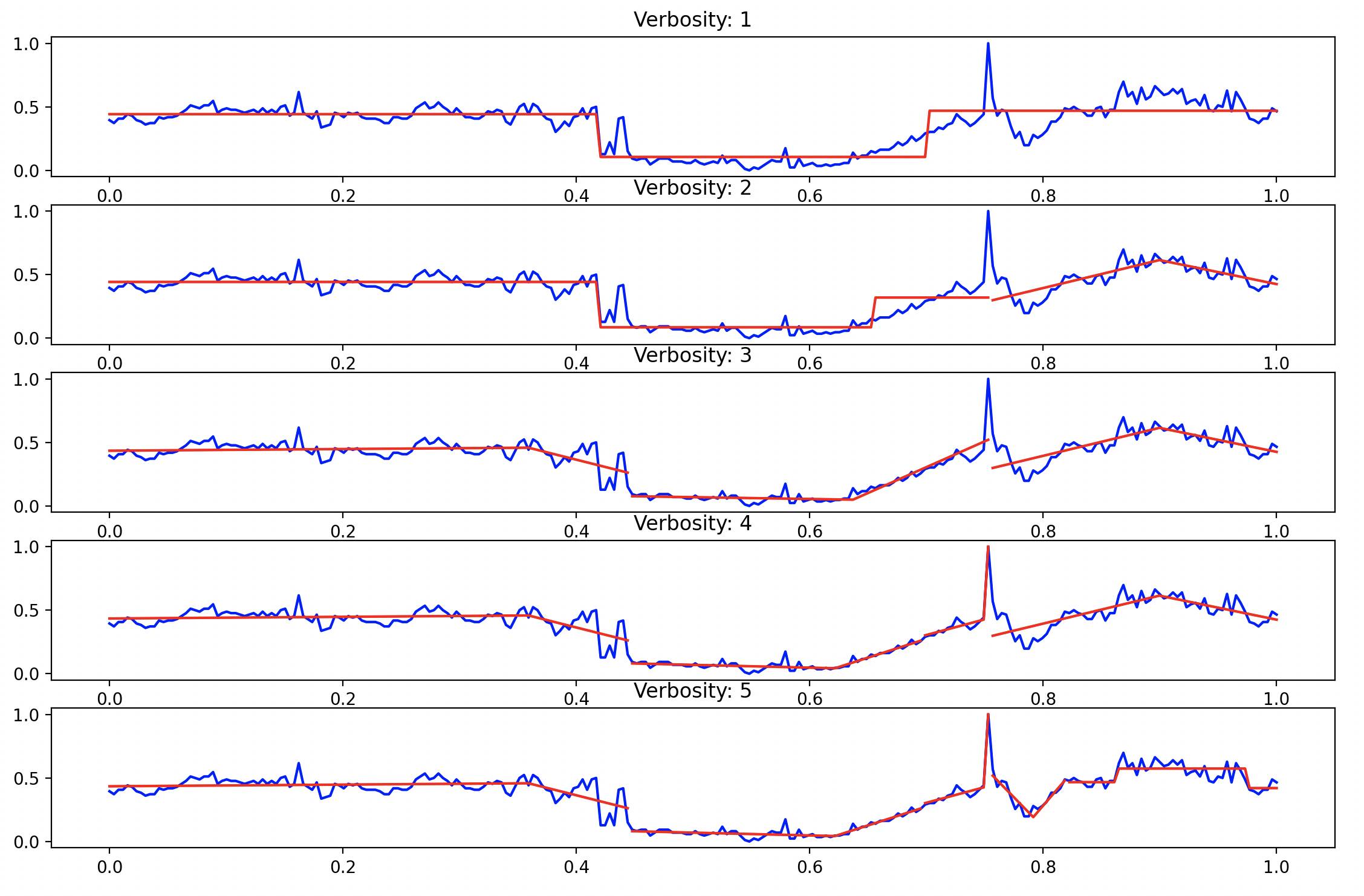}
\caption{Example of optimal descriptions for verbosity 1--5}
\label{fig:verbosity}
\end{figure}

\begin{figure}[tb]
 \begin{center}
\includegraphics[width=0.33\linewidth]{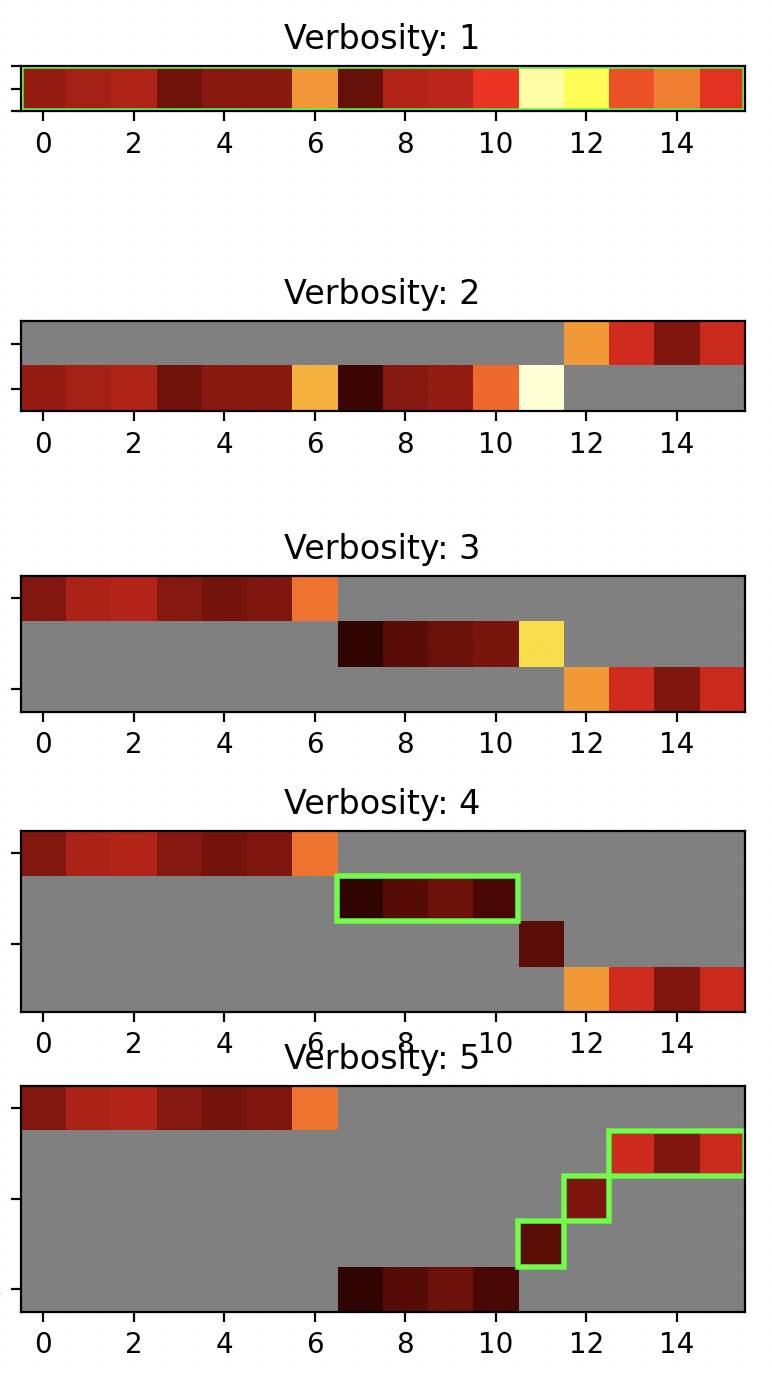}
\hspace{1ex}
\includegraphics[width=0.42\linewidth]{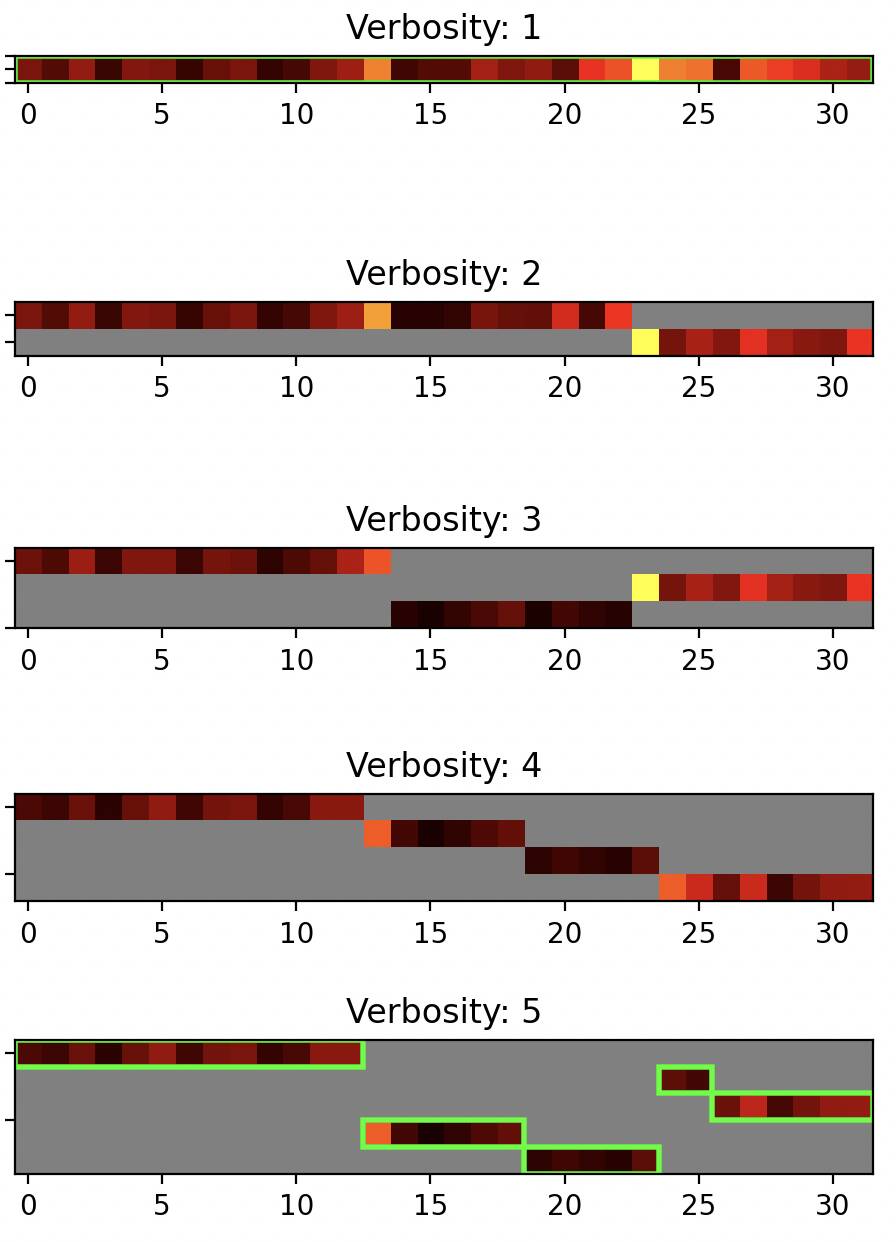}
\end{center}
\caption{Example of optimal RMSE for verbosity 1--5 (left 16 zones, right 32 zones)}
\label{fig:rmse}
\end{figure}

In Figure~\ref{fig:verbosity} it is shown, for the same dataset of Figure~\ref{fig:fitting}, the optimization results for $v\in [1..5]$. It can be noted that the red functions (descriptors) provide a better accuracy on details as more descriptors can be included. E.g., starting by a tooth function ($v=1$), the sharp peak reaching 1.0 is better approximated with $v>3$. Figure~\ref{fig:rmse} shows in a black-red-yellow palette the RMSE for each zone (on the left, 16 zones, $\ell=4$) and for each $v$. It can be noted how the error is significantly reduced by higher verbosity.

We believe that a fluent and effective description should find a balance between general synthesis and offered details.
Details should be selected in order to avoid redundant information and/or no RMSE improvement w.r.t.\ larger descriptions. Finally, relationships among descriptions (hierarchical structure) could improve the structure of the narration.
In conclusion, selecting a low (general description) and a high (details) verbosity, e.g., $1$ and $v$, would result into an unnatural flat list of sentences, with the possible outcome of repetitions and loss of interesting details.
Unfortunately, extending the previous model by these additional requirements creates a combinatorial explosion that can not be easily handled, essentially because every descriptor is a valid candidate.

\subsection{The right balance between summary and details}
We present the second optimization problem that models the creation of a summary and a set of details that describe the series only where necessary and accounting for the relationships among the other details.

We define \emph{summary} descriptors as the ones that cover the series with a minimal verbosity and a specified accuracy. This selection allows to control the minimum accuracy for cases in which the fitting functions encounter a noisy dataset and/or a distribution very different from the prototypes.
Formally, the set $S=D_s$ of summary descriptors is selected among the output of the previous optimization (sets $D_1 \dots D_v$, for verbosity up to $v$). We select the minimal verbosity level $s$ such that a threshold on maximal RMSE allowed ($max\_thr$) is satisfied by all zones: $Err(d,i)<max\_thr$ for $i\in Z_d$ and $d \in S$.

The decision variables of the optimization are the \emph{detail} descriptors. We ask for at most $v$ details that will compose the set of details of the narration. In principle, such details could be taken from $D_v$ (as mentioned above), being $|D_v|=v$. 
We prefer selecting details from $\cap_{i\leq v} D_i$, with the  introduction of some constraints among them and a cost function that accounts for accuracy improvements. The inclusion of lower verbosity levels allows us to consider coarser descriptions of details that may be suitable for narration as well as a more hierarchical narration structure that covers some overlaps. Moreover, some details from $D_v$ can be redundant and other aspects captured by lower verbosity could be introduced. Finally, comparing different verbosity configurations, the introduction of details can be balanced in terms of minimal amount of RMSE improvement as opposed to the cost of the introduction itself.

Let us start modeling the introduction of details. If two descriptions (including summary) taken from two different verbosity intersect on some zones, i.e., there is an additional more detailed description, there must be a minimal RMSE improvement on any zone in order to be willing to talk about the detail. This threshold ($min\_thr$) ensures that a new detail is introduced for a minimal improvement. Formally, $d\in D_i$ and $d'\in D_j$ with $i<j$ $\exists z. z \in Z_d\cap Z_d', Err(d,z)-Err(d',z)>min\_thr$.

Another constraint is that if a detail $d$ is selected, it can not be completely covered in all of its zones~$Z_d$ by other details of higher verbosity. In this case $d$ would be completely re-worded and thus redundant in the final textual description.

Finally, we design a maximization problem with a cost function related to the amount of RMSE improvement provided by details. Ideally, we try to select those details that improve the RMSE coverage of the summary. In particular, we focus on each zone. If a zone $i$ is covered by at least a detail, the cost is the difference between the best and the worst $Err(d,i)$ $\forall d. i\in Z_d$. We also reduce the cost by a tiny penalty $p$ proportional to the number of details that cover the same zone ($p=|{d. i\in Z_d}|$) in order to favor more distributed coverings.

Let us conclude with a consideration about the decision variables size. The current model needs only $\Sigma_{i\leq v}|D_i|=v(v-1)/2$ candidates. We believe that exploiting this hierarchy provided by verbosity levels is rather beneficial for an efficient selection of most relevant descriptors. If the two optimization problems were joined, the decision variables would become in the order of thousands, depending on the number of zones and function prototypes.

\subsection{Towards natural language}
We present now the final processing performed by the last ASP program. The goal is to create a narration structure that orders and relates the descriptions computed by the last optimization and to retrieve qualitative properties associated to descriptions. This ASP program allows a single Answer Set and it does not require any optimization. This crucial step allows to clearly define the properties and rules in use, to deliver a transparent program.

The paragraph structure is retrieved by computing lexicographic ordering of summary and details descriptions, based on zones coverage. A {\tt narration} predicate enumerates properties associated to a description. In particular it stores description location inside the narration order, grammar connective relation with next description (immediate/separated), inclusion relation with a description from lower verbosity and specific qualitative properties. 

The larger part of this ASP code is devoted to the conversion of a description into a set of qualitative properties (accompanied by some quantitative measures). Each function prototype requires a dedicated ASP fragment that produces a {\tt shape} property that describes the type of function in relation to its parameters. Moreover, it classifies qualitatively the extension of the shape. 

\begin{figure}[ht]
{\footnotesize \centering
\begin{minipage}{0.34\textwidth}
\includegraphics[width=1.00\linewidth]{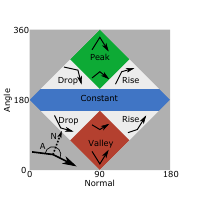}
\end{minipage}
\hfill
\begin{minipage}{0.65\textwidth}
\begin{verbatim}
select_bilin_type(ID,Angle,Normal,valley,Strength):-
      angle(Angle), angle2(Normal),  sol_desc(ID,_),
      |Angle-180|>=30,                 %%% not too flat
      Angle<180,       %%% there is a second order zero
      Normal+Angle/2 < 170,           %%% valley region
      10 < Normal+Angle/2,            %%% valley region
      adj_smooth_sharp(Strength,180-Angle,180).

adj_smooth_sharp(very_smooth,V,Max):-   V*6<Max.
adj_smooth_sharp(smooth,V,Max):-        V*6>=1*Max, V*6<2*Max.
adj_smooth_sharp(rather_smooth,V,Max):- V*6>=2*Max, V*6<3*Max.
...
\end{verbatim}
\end{minipage}
}\caption{Example of mapping parameters (normal and angle) to type of curve}
\label{fig:mapping}
\end{figure}

As an example, let us show in Figure~\ref{fig:mapping}, on the left, the mapping of parameters for a description made of two consecutive segments into an interpreted shape. In the lower left gray corner we depict two consecutive segments, characterized by a normal $N$ angle and an aperture angle $A$. Given the constraints of the three points forming the segments, $N\in [0..180]$ degrees and $A\in [0..360]$ degrees. We report the case when the two segments are rather similar in length; other cases are approximated by a single main line. As an example, the combination $N=90$ and $A=270$ would result in a peak (green area) with an smaller angle of 90 degrees. We depict the arrangements of the two segments at various combination of the two angles. Gray areas represent impossible combinations, since one of the two segments would invert the time order along x-axis. The blue area is interpreted as rather constant shape, being $A$ similar to 180. Adverbs that report on the quality the shape (e.g., rather constant, moderately constant, etc.) can be generated and the choice of intensity can be modulated by distances from optimum. 
Red and green areas represent valleys and peaks, with a rather symmetric normal. White areas represent drop and rise lines, with different flavours. 
Also, qualitative amount of shape extension along y axis is associated to an ordered list of adjectives (e.g., very mild, mild, steep, very steep) that are selected by quantizing the measures and mapping them on the list. In the same Figure, on the right, we report a fragment of ASP encoding for valley detection with conversion of the numerical quantification of its depth to a qualitative description ({\tt Strength} in the last atom in the body of the first rule).

This \emph{modus operandi}, even if crafted for each function prototype, can result in a robust general library. From the explainability point of view, the error introduced from translating a description (still a mathematical function with associated RMSE) to a qualitative narration, even if not easily measurable, it can be manually assessed by ASP code analysis and conversion maps like the one shown in Figure~\ref{fig:mapping}.

\subsection{Narration of descriptions}
Once the narration structure is output by  last ASP program, the facts are ready to be embedded in a text generation. 

As a proof of concept, we used a Python procedure that follows the structure and merges strings, to show that the only problem left is the actual string processing. Connectives are handled according to the information received and notably, no processing of relevant adjectives and nouns is performed anymore. This is to show that with minimal imperative intervention is possible to produce the text. A more fluent version could be reached if grammar rules, variants, style rules and synonyms would be implemented in a more general ASP program. This is planned as future work.

\begin{figure}[ht]
\begin{center}
 \includegraphics[width=1\linewidth]{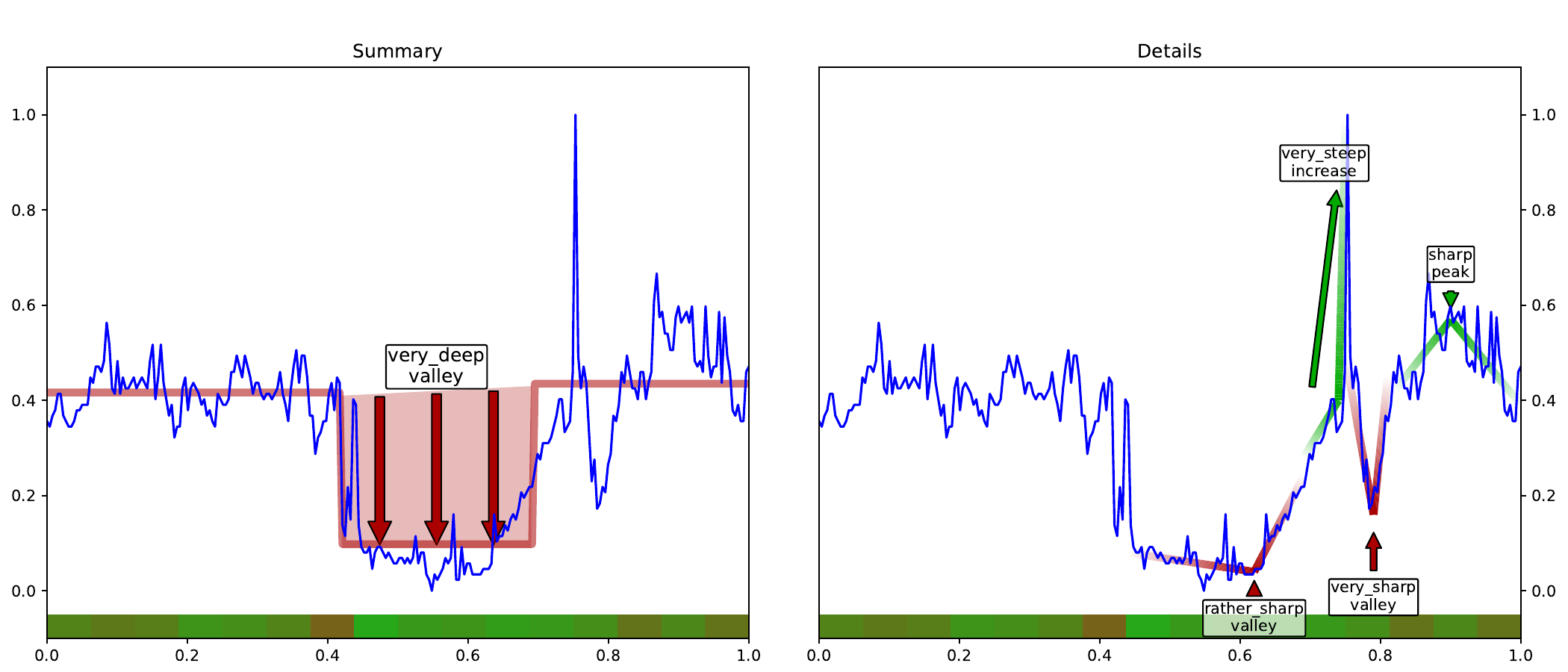}
\end{center}
\caption{Example of enriched chart. On the left the Summary and on the right the Details}
\label{fig:enriched}
\end{figure}

Another positive side effect of our processing is the capability of plotting an enriched and summarized version of the series, based on the selected descriptions, as shown in Figure~\ref{fig:enriched}. On the left we depict the summary function (in this case a valley) and on the right we highlight some peaks and valleys. On the bottom of the figures there is a red-green colored bar that reports the relative errors of each zone (16 for this example), where the red value corresponds to $max\_thr$ (0.15 in this examples).

The text associated to the Figure \ref{fig:enriched} is:

\medskip

{\footnotesize
~~~~\begin{minipage}{0.90\textwidth}
{\tt
In general, the series presents a very deep valley reaching an average of 0.09 between 0.42 and 0.69 out of a general average of 0.42 among the whole dataset. In detail, a rather sharp valley reaching a value of 0.24 occurs at 0.69; followed by a very steep increase reaching a value of 0.99 at 0.75; then by a very sharp valley reaching a value of 0.45 at 0.81; and finally by a sharp peak reaching a value of 0.39 at 1.0.   }
\end{minipage}
}

\medskip

Notice that, as depicted  in Figure~\ref{fig:enriched}, in processing the series we used a normalized scale between 0 and 1 for the x axis. A further improvement could map positions to timestamps for a more fluent processing of ranges and positions in time.

\section{Implementation}
In implementing our tool, we took advantage of Potassco Clingo Python's API. The  process is started by a Python script that downloads a tagged time series from Google Trends service.

The main program consists of a Python configurator that sends the descriptor facts to the Control object for grounding jointly with the first optimization ASP program. For the control structure of the first optimization program, we opted for an explicit branch and bound inner loop, since we want to recover optimal covering at different verbosity levels. As soon as a solution is found, the verbosity value is incremented by means of an {\tt external} feature in the outer loop. 

The second optimization and description structured are located on the same ASP program, while each function prototype conversion is stored on a separate ASP file.

Results are plotted with Matplotlib and enriching visual description are computed directly from the description structure output by the program. 

The code is available at \url{ahead-lab.unipr.it/files-for-iclp2023/}

\begin{figure}[tbh!]
\begin{center}
\includegraphics[width=.48\linewidth]{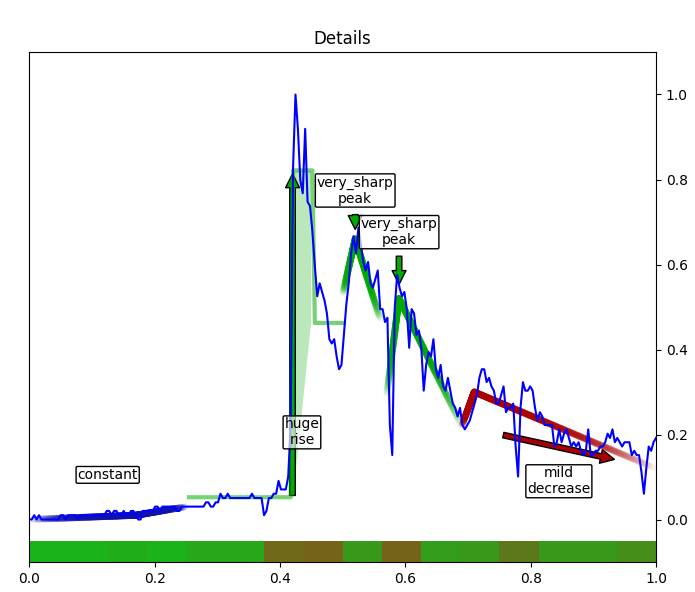}
 \hspace{1ex}
\includegraphics[width=.48\linewidth]{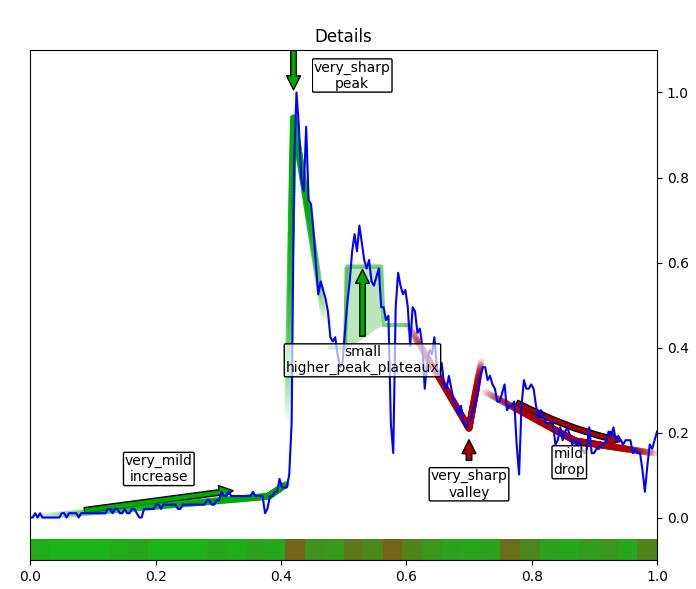}
\\~~~
{\includegraphics[width=.34\linewidth]{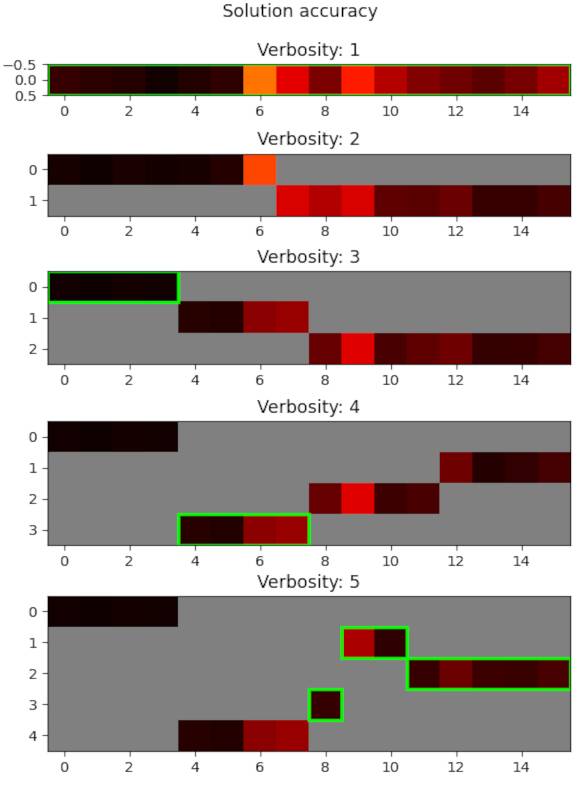}}
 \hspace{5ex}~~~~~~~
{\includegraphics[width=.45\linewidth]{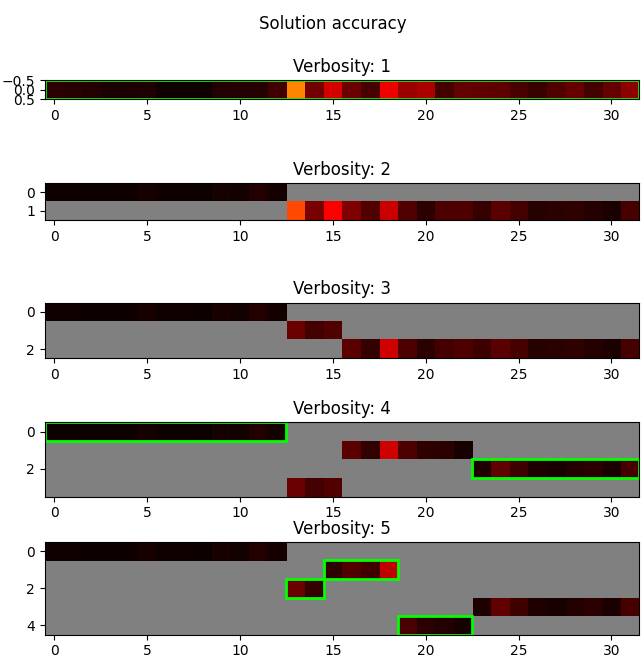}}
\end{center}
\caption{\label{fig:zonecomparison}Comparison between 16 zones (left) and 32 zones (right). On top details on the enriched plot and at bottom the heatmaps with accuracy of descriptor candidates}
\end{figure}

\section{Results}
Let us present some results and discuss the impact of some parameters changes. Defaults parameters are $max\_thr=0.15$ and $min\_thr=0.02$.

\subsection{Impact of zones on accuracy}
The number of zones clearly affects the quality of fitting functions as well as the number of descriptors. 
We compare the results in terms of overall accuracy of the narration (the sum of best contributions for each zone position, as optimized by the second program). We compare 16 vs 32 zones, that give rise to 408 and 1584 descriptors respectively (we used 3 functions prototypes).

Starting with our running example of ``Concert'' series, we can see on Figure~\ref{fig:rmse} the \emph{heatmaps} for 16 (left) and 32 (right) zones as results of the first optimization. In general colors for 32 zones are darker, meaning that descriptors are more accurate. Measuring the global RMSE (the sum for each position of the best error covered by any selected description) gives a metric to judge the accuracy of the output. Recall that the optimization goal included a penalty for overlapped descriptions. Therefore, global RMSE could be improved with a cost of more awkward narration. Nevertheless, penalties are rather low and therefore global RMSE is not too distant from the optimization cost.

Solution with 32 zones allows a selection of better descriptors and the ones selected by second optimization improve the global RMSE (0.041 with 16 zones and 0.036 with 32 zones). The global RMSE for 8 zones is 0.048, which is still acceptable, given the ability of descriptors to find optimal fitting inside the zone. However, when data has interesting patterns across zones, the ability of capturing them is lost if too few zones are used. We found that a good compromise could be 16 zones.

From a computational time point of view the complete program wall time, including function fitting (80\% of the time) is a couple of seconds for 8 zones, around 30 seconds for 16 zones and around 2 minutes for 32 zones on a standard laptop.

Let us present a second test case where the ``microsoft teams'' search term was retrieved on Google Trends. In this case global RMSE for verbosity 5 is 0.043 (8 zones), 0.030 (16 zones) and 0.024 (32 zones). 
In Figure~\ref{fig:zonecomparison} we depict on top the two detail results for 16 zones (left) and 32 zones (right). It can be noticed that more accurate zones allow to better describe the last two features (valley and drop on the right) rather than corresponding a mild decrease (left). In order to maintain the number of descriptions equal to 5, there is an introduction of a small peak plateau (left) in place of two sharp peaks (right). The overall RMSE is improved, as depicted on the green-red bar at the bottom of the chart. 
The two heatmaps at the bottom of the figure show the accuracy with a black-red-yellow scale depending on the zones. The  descriptors selected by the second optimization are colored  green.

\subsection{Verbosity comparison and text generation}

We show here the impact of verbosity, which is the most influential control for sentence summary. 

\begin{figure}[htb!]
\begin{center}
 \includegraphics[width=.48\linewidth]{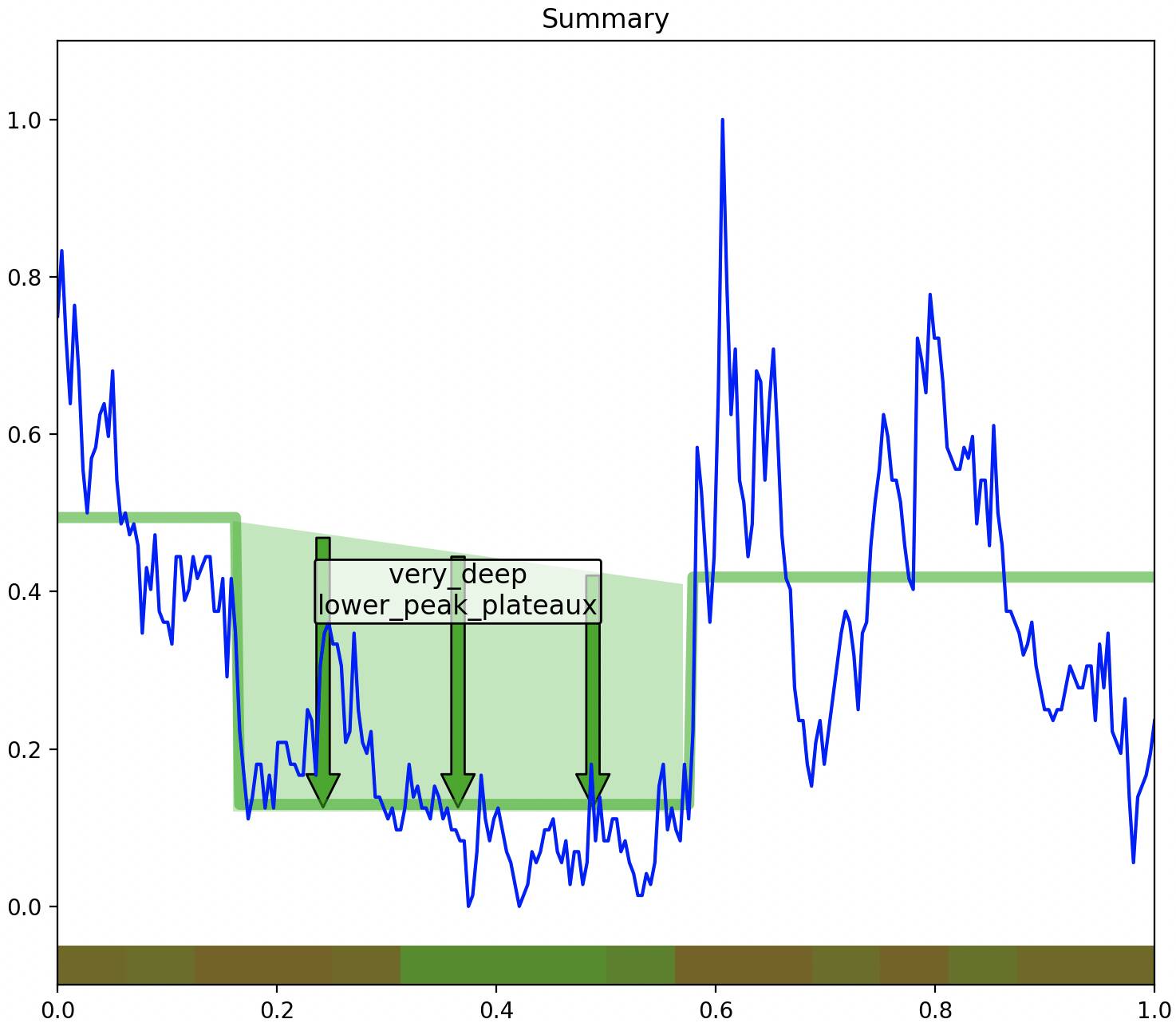}
 \hspace{1ex}
\includegraphics[width=.49\linewidth]{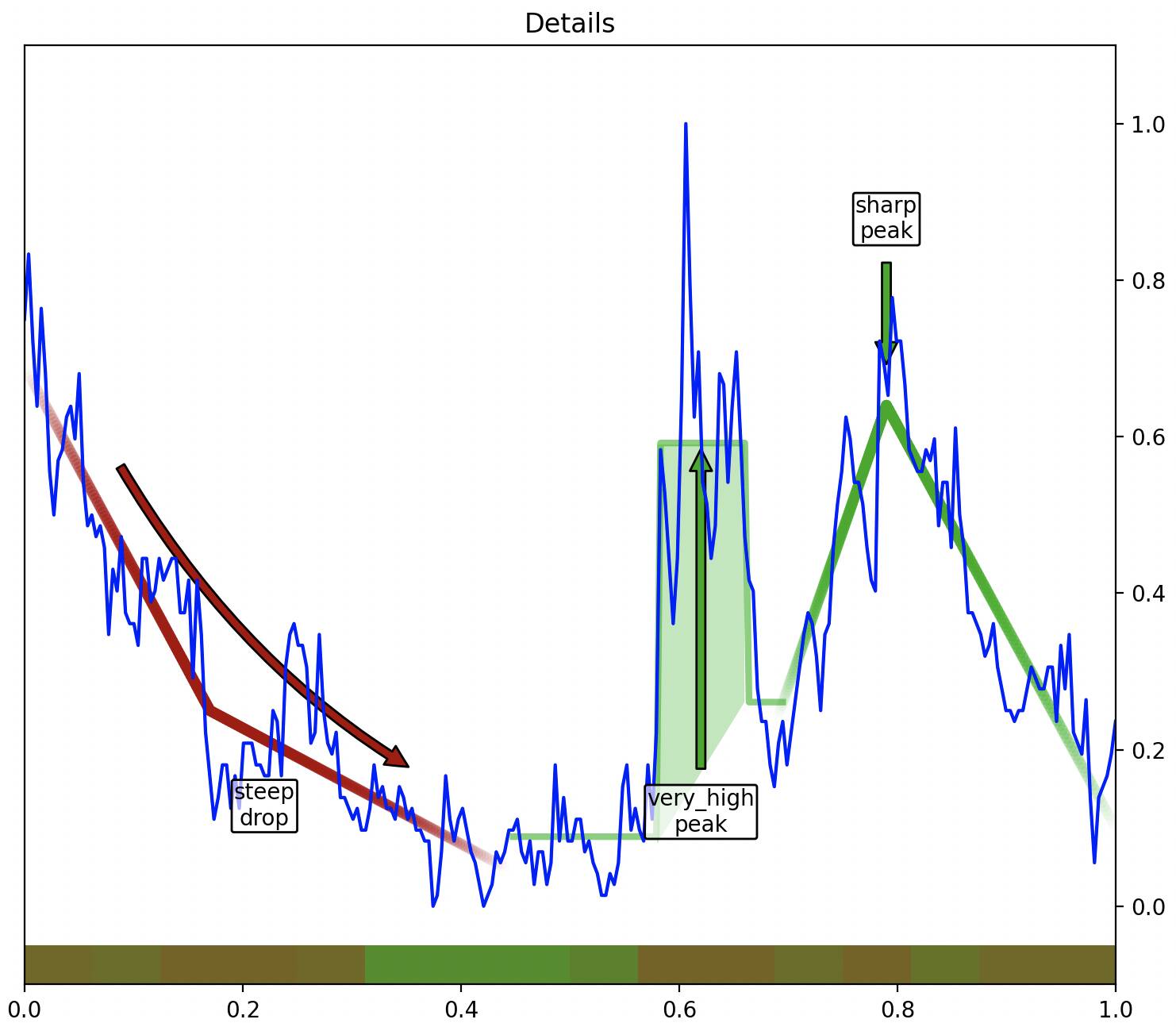}
\\
~~ \includegraphics[width=.48\linewidth]{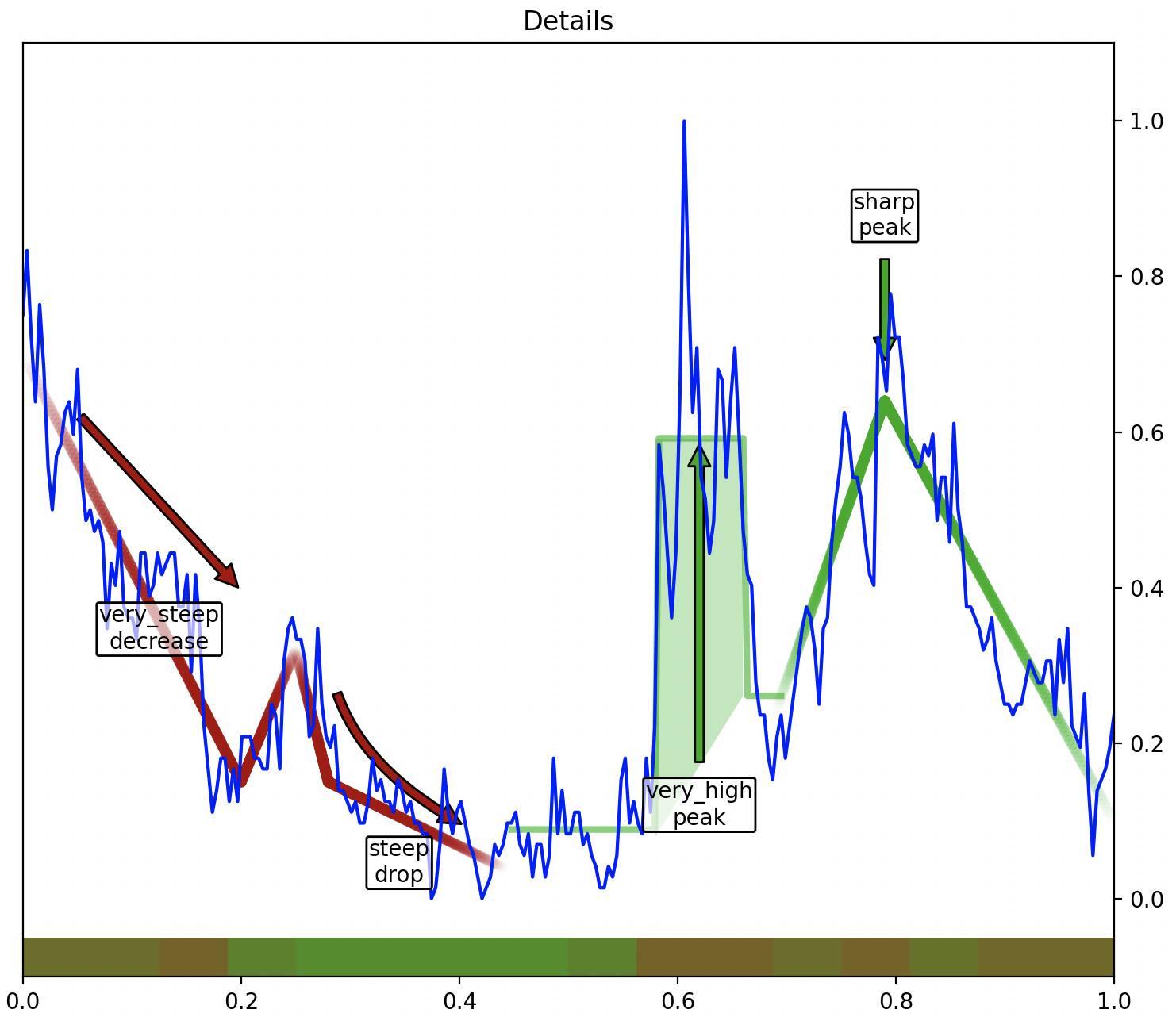}
 \hspace{0.4ex}
\includegraphics[width=.48\linewidth]{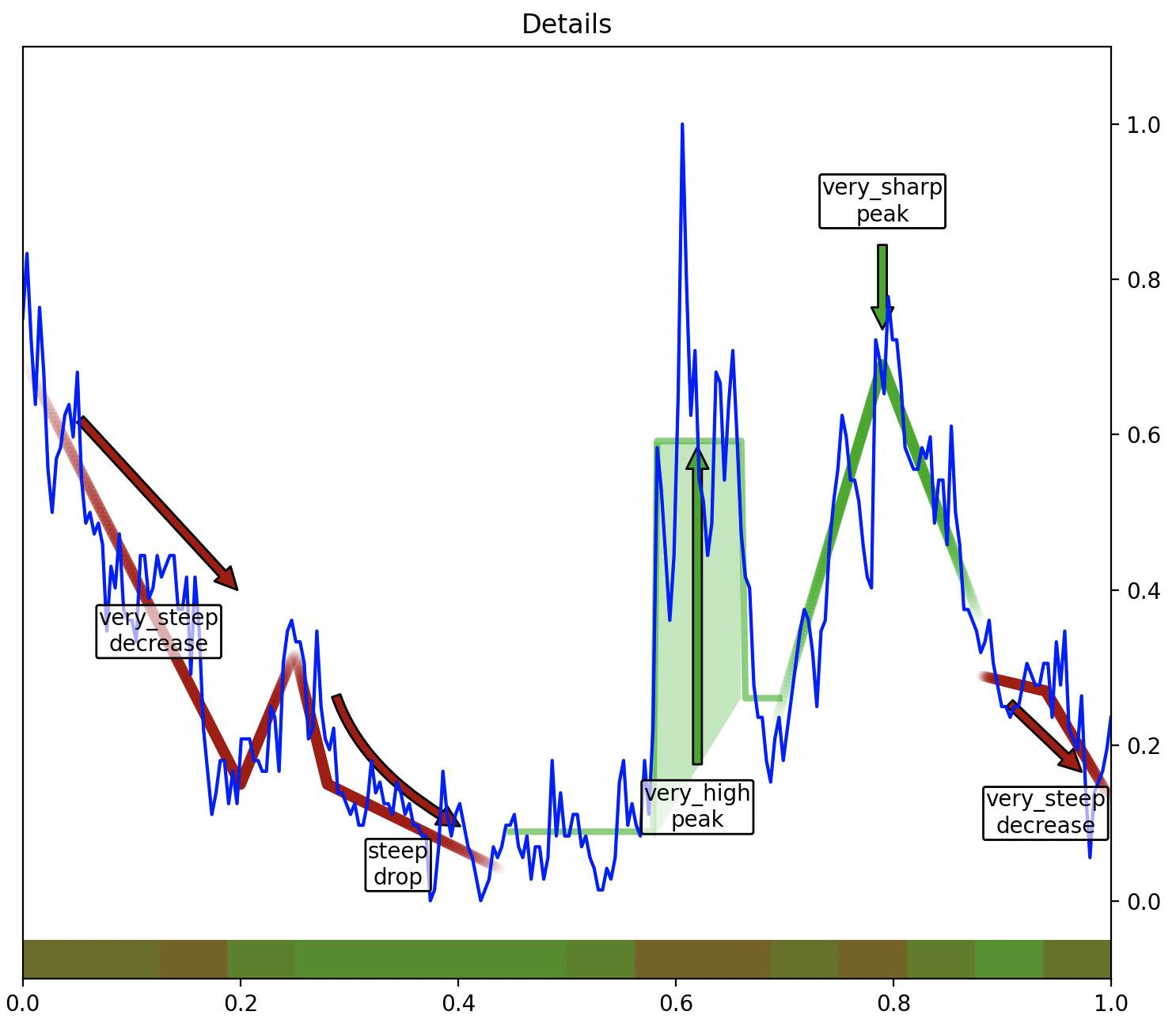}

\end{center}
\caption{Enriched charts with different verbosity (3 = top right, 4 = bottom left, 5 = bottom right). At top left the summary, shared by all solutions.}
\label{fig:verbcomparison}
\end{figure}

We setup another test case with the series obtained with the search term ``blockchain'' from Google Trends (last 5 years) with 16 zones. We launched the program with verbosity 3, 4, and 5.
Figure~\ref{fig:verbcomparison} depicts the corresponding enriched plots. On the top left the summary, which is common to all solutions, presents the main description (a deep plateau). On the top left we show verbosity 3 with the presence of an initial drop, a narrow peak and a sharp peak. It is interesting to note on bottom left (verbosity 4) that the initial drop is better approximated by a sequence of two steep drops. At their intersection a new peak is determined. As future work we plan to implement the abstraction of consecutive functions, which could result in a higher level identification of other patterns as post processing. Finally, at bottom right the introduction of another description captures the last emerging peak at the end of the series.

Let us report here the text generation from the selected descriptors. 

\medskip

{\footnotesize
\begin{minipage}{0.90\textwidth}
{\bf Verbosity~3}:
{\tt
In general, the series presents a very deep lower peak plateau reaching an average of 0.12 between 0.16 and 0.57 out of a general average of 0.45 among the whole dataset. In detail, a steep drop reaching a value of 0.05 occurs at 0.44; followed by a very high peak reaching an average of 0.59 between 0.58 and 0.66 out of a general average of 0.17 between 0.44 and 0.69; and finally by a sharp peak reaching a value of 0.1 at 1.0.
}
\end{minipage}

\medskip

\begin{minipage}{0.90\textwidth}
{\bf Verbosity~4}:
{\tt
In general, the series presents a very deep lower peak plateau reaching an average of 0.12 between 0.16 and 0.57 out of a general average of 0.45 among the whole dataset. In detail, a very steep decrease reaching a value of 0.32 occurs at 0.25; followed by a steep drop reaching a value of 0.04 at 0.44; then by a very high peak reaching an average of 0.59 between 0.58 and 0.66 out of a general average of 0.17 between 0.44 and 0.69; and finally by a sharp peak reaching a value of 0.1 at 1.0.
}
\end{minipage}

\medskip

\begin{minipage}{0.90\textwidth}
{\bf Verbosity~5}:
{\tt
In general, the series presents a very deep lower peak plateau reaching an average of 0.12 between 0.16 and 0.57 out of a general average of 0.45 among the whole dataset. In detail, a very steep decrease reaching a value of 0.32 occurs at 0.25; followed by a steep drop reaching a value of 0.04 at 0.44; then by a very high peak reaching an average of 0.59 between 0.58 and 0.66 out of a general average of 0.17 between 0.44 and 0.69; a very sharp peak reaching a value of 0.36 at 0.88; and finally by a very steep decrease reaching a value of 0.13 at 1.0.
}
\end{minipage}
}

\subsection{Conclusion}

As future work we plan to optimize the Python fitting pre-processing, being the most time demanding (several seconds). The goal is to optimize the system to deliver almost real time answers. We also plan to develop a comprehensive interpretation of consecutive descriptions, with the goal of inferring additional descriptions. In this case, some higher-level description could highlight interesting information that emerges, e.g., if two sharp peaks are adjacent, a valley is probably in between. Another ASP program could take care of the translation of the narration structure into natural language, porting some Logic Programming ideas already appeared in the literature. In the results, we showed a simple translation to natural language in Python, but a richer rule-based model (e.g., a DCG in Prolog) could handle the variants and nuances of a natural language in a more elegant and effective form. 

In general, we plan to move from single time series and to extend the work in order to handle multiple and synchronized time series. Such analysis could be extended to high level interpretation of multiple source events that could be described by an ontology. This could result in critical systems monitoring (e.g., in healthcare), where explainability is fundamental. 
Moreover, comparisons of series as well as comparisons of multidimensional data are other domains for future investigation. 

In conclusion, the paper presented an explainable methodology that produces a natural language text, coming from a structured organization of descriptions, as result of the interpretation of a time series. Reasoning about descriptions is ruled by an ASP core (transparent and domain-independent): choices about \emph{what to say} depend on an optimization problem that is controlled by an ASP concise and transparent program; the definition of \emph{how to say} is handled by an ASP program that organizes descriptions into a narration structure, assigns qualitative evaluations and is able to abstract low level features into more general ones. The output is associated to a RMSE measure on accuracy of each sentence; the conversion from descriptors to qualitative interpretation is described by means of a logic program; full control on minimal and maximal errors is provided; verbosity is a key input parameter that models the whole process; the narration structure is handled by a logic program and its description can be easily converted to natural language and to enriched graphs.

\bibliographystyle{eptcsini}
\bibliography{DaDoF.bib}
\end{document}